\documentclass{article}
\usepackage{ijcai26}

\usepackage{times}
\usepackage{soul}
\usepackage{url}
\usepackage[hidelinks]{hyperref}
\usepackage[utf8]{inputenc}
\usepackage[small]{caption}
\usepackage{graphicx}
\usepackage{amsmath}
\usepackage{amsthm}
\usepackage{booktabs}
\usepackage{algorithm}
\usepackage{algorithmic}

\usepackage[switch]{lineno}

\usepackage{amssymb}
\usepackage{inconsolata}
\usepackage{wasysym}
\usepackage{pifont}

\usepackage{multirow}
\usepackage{makecell}

\usepackage[table]{xcolor}
\definecolor{tabhead}{RGB}{215, 227, 240}
\definecolor{tabrow}{RGB}{236, 240, 245}
\definecolor{incRed}{RGB}{219, 85, 92}
\definecolor{decBlue}{RGB}{58, 140, 190}
\definecolor{tickGreen}{RGB}{46, 139, 87}
\definecolor{crossRed}{RGB}{192, 57, 43}
\definecolor{partOrange}{RGB}{230, 126, 34}

\newcommand{\cmark}{\textcolor{tickGreen}{\CheckedBox}}
\newcommand{\xmark}{\textcolor{crossRed}{\XBox}}
\newcommand{\hmark}{{\color{partOrange}\scalebox{0.83}{$\boxminus$}}}
\newcommand{\inc}[1]{\textsuperscript{\textcolor{incRed}{\raisebox{0.1ex}{$\uparrow$}#1}}}
\newcommand{\dec}[1]{\textsuperscript{\textcolor{decBlue}{\raisebox{0.1ex}{$\downarrow$}#1}}}

\newcommand{\correspondingmark}{\textsuperscript{\dag}}

\title{MetaGen: Self-Evolving Roles and Topologies for Multi-Agent LLM Reasoning}

\author{
Yimeng Wang$^1$\textsuperscript{*}
\and
Jiaxing Zhao$^2$\textsuperscript{*}\and
Hongbin Xie$^2$\and
Hexing Ma$^1$\\
Yuzhen Lei$^1$\and
Shuangxue Liu$^1$\and
Xuan Song$^{1,2}$\correspondingmark\and
Zichen Zhang$^3$\And
Haoran Zhang$^3$\correspondingmark
\affiliations
$^1$School of Artificial Intelligence, Jilin University\\
$^2$Department of Computer Science and Engineering, Southern University of Science and Technology\\
$^3$School of Urban Planning and Design, Peking University\\
\emails
\{yimeng24, jiaxing25, hxma24, leiyz25, sxliu25\}@mails.jlu.edu.cn,
12131108@mail.sustech.edu.cn,
songxuan@jlu.edu.cn,
\{zhangzc9752, h.zhang\}@pku.edu.cn
}

\begin{document}
\maketitle

\begingroup
\renewcommand\thefootnote{*} % 将当前脚注标记设置为星号
\footnotetext{Equal contribution.} % 星号对应的脚注内容

\renewcommand\thefootnote{\dag} % 将当前脚注标记设置为剑号
\footnotetext{Corresponding author.} % 剑号对应的脚注内容
\endgroup

\begin{abstract}
Large language models are increasingly deployed as multi-agent systems, where specialized roles communicate and collaborate through structured interactions to solve complex tasks that often exceed the capacity of a single agent. However, most existing systems still rely on a fixed role library and an execution-frozen interaction topology, a rigid design choice that frequently leads to task mismatch, prevents timely adaptation when new evidence emerges during reasoning, and further inflates inference cost. We introduce \textbf{MetaGen}, a training-free framework that adapts both the role space and the collaboration topology at inference time, without updating base model weights. MetaGen generates and rewrites query-conditioned role specifications to maintain a controllable dynamic role pool, then instantiates a constrained execution graph around a minimal backbone. During execution, it iteratively updates role prompts and adjusts structural decisions using lightweight feedback signals. Experiments on code generation and multi-step reasoning benchmarks show that MetaGen improves the accuracy and cost tradeoff over strong multi-agent baselines.
\end{abstract}

\section{Introduction}
\label{sec:intro}

Large language models (LLMs) are rapidly evolving from single-turn conversational responders into general-purpose problem solvers that can plan, critique, write code, and interact with external tools\cite{debate,shinn2023reflexion}.
A natural next step is to organize multiple LLM instances into Multi-Agent Systems (MAS)\cite{li2023camel,autogen,tang2024medagents,chen2024reconcile,liu2025rcr}, where specialized roles collaborate to decompose complex tasks and cross-check intermediate conclusions.
Recent systems have shown that role-playing and structured collaboration can substantially outperform single-agent prompting on reasoning, tool use, and software engineering workflows \cite{chen2024agentverse,metagpt}.
At the same time, prompting paradigms such as debate, reflection, and search-based reasoning point to a broader lesson: the interaction structure---who speaks, what is produced, and how signals are aggregated---can be as influential as the base model itself \cite{debate,yao2022react,besta2024got}.

\begin{figure}[t]
\includegraphics[width=\columnwidth]{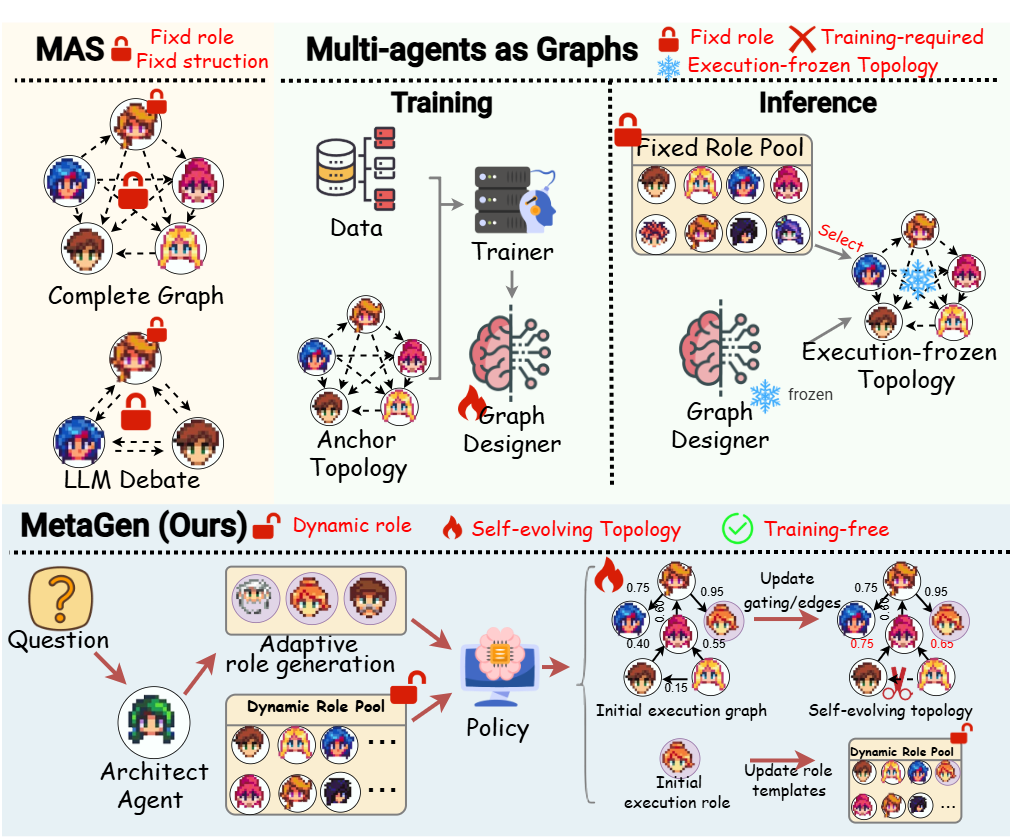} \caption{Overview and positioning of MetaGen. Unlike fixed-role/fixed-topology multi-agent systems and training-based topology designers with execution-frozen graphs, MetaGen enables training-free, query-conditioned role generation and self-evolving topology adjustment entirely at inference time.} \label{fig:teaser}
\end{figure}

Despite this progress, many deployed MAS still follow a rigid preset design.
Developers typically maintain a fixed role pool (e.g., planner/solver/verifier) and hard-code an execution-frozen message-passing protocol~\cite{qian2024scaling} (e.g., chain, star, or fully connected chat).
Such rigidity leads to three recurring issues.
First, \emph{task mismatch} arises because task granularity, tool preferences, and error modes vary widely, while a fixed role set is often brittle under distribution shift.
Second, \emph{structural closure} occurs when a topology determined once cannot be revised mid-run in response to new evidence or contradictions.
Third, \emph{cost } suffers because tailoring prompts and interaction structures to each task requires manual engineering.

An increasing line of work treats collaboration topology as a key lever and seeks to automate it \cite{yue2025masrouter,zhang2025multi}.
Graph-based views model agents as nodes and communications as directed edges, enabling orchestration search, pruning, and topology optimization~\cite{gptswarm,dylan,zhang2024cut}.
Recent topology designers learn or generate task-adaptive graphs, for example by predicting edges with graph models~\cite{g-designer} or autoregressively generating a team and its links from a query~\cite{arg-designer}.
While these approaches reduce manual engineering, two assumptions remain common at inference time: roles are drawn from a pre-defined library, and the instance-specific graph is typically frozen once execution begins. These observations motivate a central question: can an MAS generate the roles it needs and update its collaboration structure during inference, while keeping cost bounded?

We present \textbf{MetaGen}, a training-free framework that adapts both the role space and the collaboration topology at test time (Figure~\ref{fig:teaser}).
MetaGen introduces an Architect that synthesizes and revises query-conditioned role specifications to form a controllable dynamic role pool.
It then constructs an initial execution graph around a minimal backbone and iteratively updates role prompts and structural decisions using lightweight feedback signals, without modifying backbone weights.
To prevent unrestricted chatter, MetaGen enforces explicit controls, including schema/validity checks for generated roles, constrained graph construction, selective activation and edge gating, and cost-aware stopping.

MetaGen is designed to be effective and inspectable.
It logs generated roles, selected participants, structural edits, and the feedback that triggers each update, supporting reproducibility and diagnosis beyond ad hoc orchestration.
This combination of dynamic roles, inference-time evolution, and structured control targets the core limitations of rigid MAS while retaining the engineering advantages of graph-based collaboration. In summary, our contributions are:

\begin{itemize}
    \item We propose \textbf{MetaGen}, a training-free framework that improves multi-agent collaboration by adapting role specifications and communication topology during inference.
    \item We introduce \emph{query-conditioned} role generation and revision with lightweight validity constraints, yielding a controllable dynamic role pool.
    \item We develop an inference-time evolution loop that updates prompts and structural decisions under explicit constraints to bound cost and maintain auditability.
    \item Extensive experiments demonstrate that MetaGen consistently improves the accuracy--cost trade-off over competitive multi-agent baselines, and ablations confirm the complementary benefits of dynamic roles, within-instance refinement, and cross-instance accumulation.
\end{itemize}

\begin{figure*}[t]
  \centering
 \includegraphics[width=\textwidth]{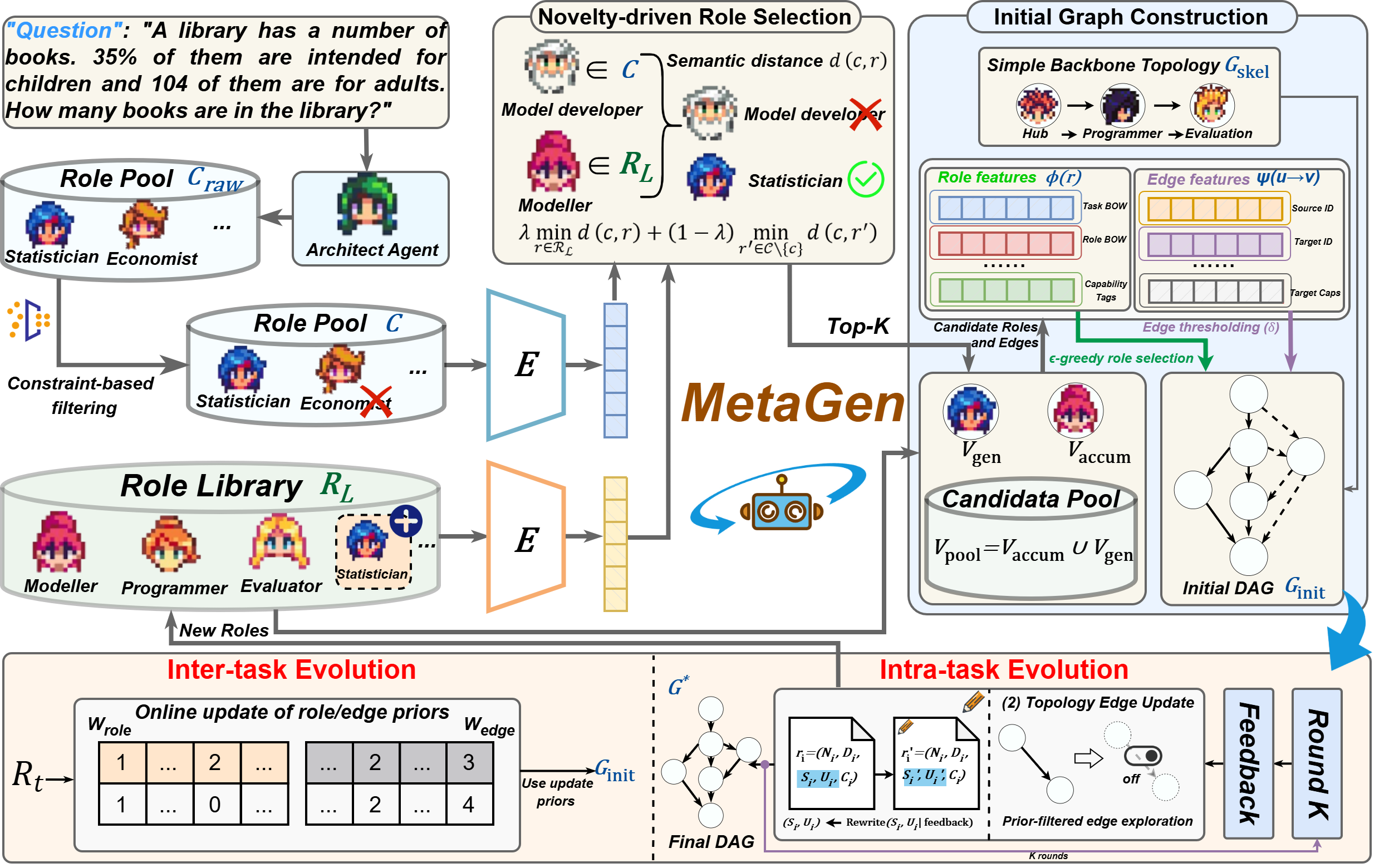}
  \caption{\textbf{MetaGen framework overview.}
    Given a query, an Architect generates and filters candidate roles, then performs novelty-driven role selection and hybrid graph initialization to form an initial DAG $G_{\text{init}}$.
    MetaGen supports intra-task evolution by updating role prompts and structure using execution feedback, and inter-task evolution by accumulating cross-instance priors and solidifying verified roles for future reuse.}
  \label{fig:pipeline}
\end{figure*}

\section{Related Work}
\subsection{Multi-agent collaboration with LLMs.}
A growing body of work solves complex tasks via LLM-based multi-agent collaboration~\cite{akata2025playing,guo2024large,zhao2024longagent,hao2025chatllm}, where multiple agents exchange intermediate results to reduce single-agent blind spots and improve reliability.
Common paradigms include discussion-style coordination that aggregates diverse perspectives and iteratively refines candidate solutions~\cite{saha2024branch}, and debate-style protocols that surface contradictions and encourage self-correction through adversarial critique~\cite{xiong2023examining}.
Another line of work emphasizes specialization by assigning distinct roles (e.g., planner, executor, verifier) and organizing them into hierarchical pipelines for decomposition and verification~\cite{zhang2025planning}.
Beyond pipelines, structured collaboration patterns such as chain and star orchestration~\cite{metagpt,chatdev,zhou2023large} and richer tree/graph-structured interaction~\cite{zhang2024coa,zhao2024longagent,soa,qian2024scaling,gptswarm,zhao2025connecting} have been adopted to better capture information dependencies and multi-step reasoning. Despite these advances, most systems assume a pre-defined role pool and adopt an interaction topology that is fixed or drawn from a small set of templates, remaining largely \emph{execution-frozen} once inference begins.
As a result, the collaboration strategy often cannot be tailored to instance-specific needs, and mismatched roles or redundant interactions can waste computation and hinder robustness under distribution shift.

\subsection{Multi-Agents as Graphs}
Graph-structured formulations are a natural fit for multi-agent collaboration, as they explicitly model information dependencies and interaction constraints among agents\cite{zhang2025multi}. Representative systems such as MacNet~\cite{qian2024scaling} and GPTSwarm~\cite{gptswarm} treat agent interaction as an optimizable graph. Recent work further moves toward \emph{dynamic} topology construction. DyLAN~\cite{dylan} selects and routes agents by filtering an initially large set based on instance-level importance signals, G-Designer~\cite{g-designer} synthesizes communication graphs with a learned generator to adapt connectivity patterns, and ARG-Designer~\cite{arg-designer} autoregressively constructs agent groups together with their links under task conditioning. Unlike topology-centric designers that primarily generate interaction graphs over fixed/retrieved roles, MetaGen treats both role specifications and topology as editable inference-time objects, enabling coupled intra-instance refinement and cross-instance accumulation.

\begin{table*}[!t]
\centering
\scriptsize
\setlength{\tabcolsep}{4.5pt}
\begingroup
\setlength{\aboverulesep}{0.2ex}
\setlength{\belowrulesep}{0.2ex}

\resizebox{\textwidth}{!}{%
% 修改点：将后半部分数值列改回 c (居中)，这样在有 \inc 标注时最整齐
% l: 左对齐
% c: 其余全部居中
\begin{tabular}{lccccccccc} 
\toprule
\rowcolor{tabhead}
\textbf{Method} &
\textbf{Dyn.} &
\textbf{T-free} &
\textbf{Evol.} &
\textbf{GSM8K} &
\textbf{HumanEval} &
\textbf{MMLU} &
\textbf{AQuA} &
\textbf{MNLI} &
\textbf{Average} \\
\midrule

\rowcolor{tabrow}
Vanilla &
\xmark & \cmark & \xmark &
89.3 &
65.2 &
87.1 &
69.7 &
77.6 &
77.8 \\

CoT (zero-shot) &
\xmark & \cmark  & \xmark &
93.1\inc{3.8} &
89.0\inc{23.8} &
89.5\inc{2.4} &
70.9\inc{1.2} &
82.3\inc{4.7} &
85.0 \\

\rowcolor{tabrow}
CoT (few-shot) &
\xmark & \cmark  & \xmark &
95.8\inc{6.5} &
92.1\inc{26.9} &
91.5\inc{4.4} &
84.6\inc{14.9} &
\underline{85.4}\inc{7.8} &
89.9 \\

SC (K=3) &
\xmark & \cmark  & \xmark &
94.2\inc{4.9} &
86.0\inc{20.8} &
90.8\inc{3.7} &
72.0\inc{2.3} &
83.8\inc{6.2} &
85.4 \\

\rowcolor{tabrow}
SC (K=10) &
\xmark & \cmark  & \xmark &
93.9\inc{4.6} &
84.1\inc{18.9} &
\underline{92.2}\inc{5.1} &
77.2\inc{7.5} &
84.0\inc{6.4} &
86.3 \\

\midrule
\rowcolor{tabrow}
Chain &
\xmark & \cmark & \xmark &
92.0\inc{2.7} &
90.2\inc{25.0} &
91.5\inc{4.4} &
79.1\inc{9.4} &
77.2\dec{0.4} &
86.0 \\

Star &
\xmark & \cmark & \xmark &
94.5\inc{5.2} &
89.6\inc{24.4} &
90.2\inc{3.1} &
83.5\inc{13.8} &
69.9\dec{7.7} &
85.5 \\

\rowcolor{tabrow}
Tree &
\xmark & \cmark & \xmark &
77.5\dec{11.8} &
93.9\inc{28.7} &
77.1\dec{10.0} &
83.9\inc{14.2} &
55.3\dec{22.3} &
77.5 \\

Complete Graph &
\xmark & \cmark & \xmark &
94.6\inc{5.3} &
89.0\inc{23.8} &
\underline{92.2}\inc{5.1} &
86.2\inc{16.5} &
83.3\inc{5.7} &
89.1 \\

\rowcolor{tabrow}
Random Graph &
\xmark & \cmark & \xmark &
95.4\inc{6.1} &
92.1\inc{26.9} &
91.8\inc{4.7} &
78.7\inc{9.0} &
84.2\inc{6.6} &
88.4 \\

LLM-Debate &
\xmark & \cmark & \xmark &
94.2\inc{4.9} &
89.6\inc{24.4} &
\underline{92.2}\inc{5.1} &
85.8\inc{16.1} &
79.7\inc{2.1} &
88.3 \\

\rowcolor{tabrow}
GPTSwarm &
\xmark & \xmark & \xmark &
-- &
69.6\inc{4.4} &
60.1\dec{27.0} &
-- &
-- &
64.9 \\

AFlow &
\xmark & \xmark & \hmark &
94.3\inc{5.0} &
90.9\inc{25.7} &
-- &
-- &
-- &
92.6 \\

\rowcolor{tabrow}
G-Designer &
\xmark & \xmark & \hmark &
\underline{96.3}\inc{7.0} &
\underline{94.2}\inc{29.0} &
\textbf{93.5}\inc{6.4} &
89.0\inc{9.3} &
-- &
\underline{93.3} \\

ARG-Designer &
\cmark & \xmark & \hmark &
96.1\inc{6.8} &
90.9\inc{25.7} &
89.5\inc{2.4} &
\underline{90.6}\inc{20.9} &
-- &
91.8 \\

\midrule
\rowcolor{tabrow}
\textbf{MetaGen} &
\cmark & \cmark & \cmark &
\textbf{96.4}\inc{7.1} &
\textbf{95.1}\inc{29.9} &
\textbf{93.5}\inc{6.4} &
\textbf{95.7}\inc{26.0} &
\textbf{94.8}\inc{17.2} &
\textbf{95.1} \\

\bottomrule
\end{tabular}%
}
\caption{\label{main}
Main results on five benchmarks using DeepSeek-V3.
Dyn.\ indicates whether a method uses a dynamic role pool.
T-free indicates whether it avoids training model weights for role or topology design.
Evol.\ indicates whether the interaction topology evolves during inference.
\cmark means yes.
\xmark means no.
\hmark means partial. All numbers are means over three independent runs.
}
\endgroup
\end{table*}

\section{Method}
\textbf{MetaGen} is a training-free framework for multi-agent collaboration that models role specifications and communication topology as first-class, editable entities during inference. It enables structured adaptation via query-conditioned role generation and revision, coupled with a self-evolving graph orchestration loop subject to explicit structural constraints. Under this formulation, collaboration structures are progressively refined to meet task-specific requirements, while bounding computational cost and preserving auditable interaction traces, thereby improving adaptability, reproducibility, and reasoning performance in multi-agent systems.

\subsection{Problem Formulation}
Given a task input $x$ (e.g., code generation or complex reasoning), \textbf{MetaGen} employs an \textit{Architect Agent} during inference to automatically generate a set of candidate agent roles $\{\hat{r}_i\}_{i=1}^N$, from which a directed acyclic graph (DAG) $G=(V,E)$ is constructed to model the multi-agent collaboration process.

Each node $v \in V$ corresponds to a specific agent role, and each directed edge $e=(u \rightarrow v) \in E$ represents a message flow between roles, capturing information dependencies and collaboration pathways during multi-agent reasoning. Each role $r_i$ is formally defined as a tuple:
\begin{equation}
    r_i = (N_i,\, D_i,\, S_i,\, U_i,\, C_i),
\end{equation}
where $N_i$ denotes the role name, $D_i$ denotes the semantic description of the role, $S_i$ denotes the system-level prompt template, $U_i$ denotes the user-level prompt template, and $C_i$ denotes the set of capabilities or tools available to the role.

Without updating the underlying large language model parameters, the core objective of MetaGen is to regulate the multi-agent inference process through joint optimization of role specifications and collaboration structure by minimizing the following composite objective:
\begin{equation}
\begin{aligned}
    \min_{\text{arch},\,\text{roles}} \; \mathcal{L}
    = {} & \lambda_1 \mathcal{L}_{\mathrm{acc}}(y, y^*)
         + \lambda_2 \mathcal{L}_{\mathrm{cost}}(\tau) \\
         & + \lambda_3 \mathcal{L}_{\mathrm{sparse}}(G),
\end{aligned}
\end{equation}
where $\mathcal{L}_{\mathrm{acc}}$ quantifies the prediction error between the system output $y$ and the ground-truth target $y^*$, $\mathcal{L}_{\mathrm{cost}}$ penalizes the cumulative token usage and inference latency over the reasoning trajectory $\tau$, and $\mathcal{L}_{\mathrm{sparse}}$ serves as a structural regularizer that promotes sparsity in the communication graph, improving interpretability and controllability. During inference, MetaGen does not access $y^*$; it relies on naturally observable execution signals to trigger edits and updates only lightweight selection priors from the pass/cost summary.

\subsection{Generative Role Space}
\label{sec:generative_roles}
To address task mismatch and distribution shift, MetaGen implements a dynamic \textit{Generative Role Space}. We employ an Architect Agent to synthesize a raw candidate set $\mathcal{C}_{\text{r}}$ conditioned on the query. To ensure the role space is both executable and non-redundant, we enforce a formalized two-stage validation process.

\paragraph{Constraint-Based Filtering.}
We first refine the raw generations into a valid candidate set $\mathcal{C}$ by imposing strict structural and safety constraints. Let $T(c)$ denote the prompt template of candidate $c$, and $\mathcal{W}(c)$ be its token set. We define the valid set as:
\begin{equation}
    \mathcal{C} = \left\{ c \in \mathcal{C}_{\text{r}} \mid \left( T(c) \models \Phi \right) \land \left( \mathcal{W}(c) \cap \mathcal{V}_{\text{b}} = \emptyset \right) \right\},
\end{equation}
where $\Phi$ represents the required schema (e.g., placeholders), $\models$ denotes schema satisfaction, and $\mathcal{V}_{\text{b}}$ is a set of restricted keywords.

\paragraph{Embedding-Based Diversity Gating.}
To avoid semantic redundancy, we project roles into a dense vector space to enforce diversity. Let $\mathcal{E}: \mathcal{X} \to \mathbb{R}^d$ denote a semantic encoder that maps the textual description of a role $c$ to its embedding vector $\mathbf{e}_c$:
\begin{equation}
    \mathbf{e}_c = \frac{\mathcal{E}(\text{desc}(c))}{\| \mathcal{E}(\text{desc}(c)) \|_2}.
\end{equation}
To prevent semantic redundancy, we employ an embedding-based ranking mechanism. Let $d(c, r)$ denote the semantic distance between two roles in the embedding space.

For each candidate $c \in \mathcal{C}$, we calculate a \textit{Marginal Utility Score} $S(c)$ that balances external novelty against the historical library $\mathcal{R}_L$ and internal distinctiveness relative to other candidates:
\begin{equation}
    S(c) = \lambda \min_{r \in \mathcal{R}_L} d(c, r) + (1-\lambda) \min_{r' \in \mathcal{C} \setminus \{c\}} d(c, r'),
\end{equation}
where $\lambda$ controls the trade-off weight. Finally, we construct the incremental role set $\Delta \mathcal{R}$ by selecting the top-$K$ candidates that exceed a minimum novelty threshold $\delta$:
\begin{equation}
    \Delta \mathcal{R} = \operatorname{Top}_K \left( \left\{ c \in \mathcal{C} \mid S(c) > \delta \right\} \right).
\end{equation}
This selection strategy ensures that the instantiated roles are not only valid but also semantically unique and non-redundant.

\subsection{Task-Adaptive Graph Construction}
\label{sec:graph_construction}
To balance structural regularization with semantic flexibility, we propose a hybrid graph construction strategy. It anchors reasoning to a minimal functional backbone, expands it through score-based selection over a hybrid role pool, and supports evolution at two levels: intra-task refinement within an instance and inter-task accumulation across instances.

\paragraph{Hybrid Graph Initialization.}
We first instantiate a task-type backbone chain $G_{\text{skel}}=(V_{\text{skel}},E_{\text{skel}})$ to guarantee the fundamental execution flow. For code generation, the chain is
\begin{equation}
    E_{\text{skel}}=\{(v_{\text{hub}}\!\to\!v_{\text{prog}}),\ (v_{\text{prog}}\!\to\!v_{\text{eval}})\}.
\end{equation}
Here $v_{\text{hub}}$ dispatches the request, $v_{\text{prog}}$ produces code, and $v_{\text{eval}}$ verifies it.

To handle requirements beyond the backbone, we form a hybrid candidate pool
$\mathcal{V}_{\text{pool}}=\mathcal{V}_{\text{accum}}\cup\mathcal{V}_{\text{gen}}$,
where $\mathcal{V}_{\text{accum}}$ contains accumulated generalist and previously effective roles, and $\mathcal{V}_{\text{gen}}$ contains query-conditioned roles synthesized by the Architect for the current instance.

Each candidate role $r$ is represented by a feature vector $\phi(r)$ that combines lexical cues from the role name/prompt template, capability indicators, semantic relevance to the query via $\mathcal{E}$, and optional historical statistics when available.
Each candidate directed edge $(u\!\to\!v)$ is represented by $\psi(u\!\to\!v)$, which combines endpoint features with simple structural signals and optional co-occurrence statistics.
We compute linear priority scores
\begin{equation}
    s_r=\mathbf{w}_{\text{role}}^\top \phi(r),\qquad
    s_{u\to v}=\mathbf{w}_{\text{edge}}^\top \psi(u\!\to\!v),
\end{equation}
and select a Top-$K$ committee with an $\epsilon$-greedy strategy.
Edges are added to form $G_{\text{init}}$ when their scores exceed a threshold $\delta$, subject to DAG constraints.

\paragraph{Intra-task Evolution.}
Starting from $G_{\text{init}}$, MetaGen performs lightweight within-instance refinement over multiple rounds.
We denote by $\mathcal{F}$ the \textit{feedback} collected during inference and tool execution, consisting of naturally observable signals such as runtime logs, compilation/test outcomes, format validators, and self-consistency checks.
This feedback is available without introducing additional supervision and serves as the trigger for instance-level edits.

Given $\mathcal{F}$, MetaGen applies two types of edits that operate \emph{only} on textual role specifications and a constrained subset of structural choices.
First, \emph{role prompt rewrite} targets a low-utility role whose messages are consistently unhelpful (e.g., redundant, unstable, or verbose) under the current instance.
Using the feedback traces (error messages, failed checks, or inconsistency patterns), MetaGen revises the role's system/user templates to better align the role behavior with the instance requirements.
Second, \emph{prior-filtered edge exploration} updates topology within the instance in a conservative manner.
MetaGen first filters candidate \emph{non-critical} edges using current priors and structural constraints (e.g., preserving at least one path to the exit/judge node and avoiding cycles), then selectively deactivates or swaps one edge to encourage simpler, more informative communication.
Across rounds, these edits allow the collaboration process to react to observed failure modes while keeping the execution stable and auditable.

\paragraph{Inter-task Evolution.}
While intra-task evolution adapts behavior for a single instance, MetaGen also improves future decisions by maintaining lightweight state across instances.
After an instance completes, we summarize its overall outcome into a scalar \textit{reward} that trades off success and cost,
\begin{equation}
    R=\mathbb{I}(\text{pass})-\lambda_{\text{cost}}\cdot \mathcal{C}_{\text{token}},
\end{equation}
where $\mathbb{I}(\text{pass})$ is a task-specific pass indicator produced by the evaluator and $\mathcal{C}_{\text{token}}$ is the total token usage.
We then update the parameters that govern role/edge scoring with a reward-weighted linear rule:
\begin{equation}
    \mathbf{w} \leftarrow \mathbf{w} + \eta\,R\,\mathbf{f},
    \label{eq:rw_update}
\end{equation}
where $\mathbf{f}$ is the feature vector for the decision that was used, i.e., $\mathbf{f}=\phi(r)$ for a selected role (updating $\mathbf{w}_{\text{role}}$) or $\mathbf{f}=\psi(u\!\to\!v)$ for an activated edge (updating $\mathbf{w}_{\text{edge}}$).
Intuitively, decisions that lead to successful, low-cost executions receive positive updates and become more likely under similar contexts, whereas costly or unsuccessful executions yield weaker (or negative) reinforcement.
This mechanism is deliberately lightweight: it updates only shallow priors used by the selector/wiring module and remains fully decoupled from backbone LLM weight training.

\begin{algorithm}[tb]
\caption{MetaGen: inference-time evolution of roles and topology}
\label{alg:metagen}
\textbf{Input}: task input $x$, role library $\mathcal{R}_L$, skeleton $G_{\text{skel}}$\\
\textbf{Parameter}: Top-$K$, $\epsilon$, $\delta$, $T_{\max}$, $\eta$, $\lambda_{\text{cost}}$\\
\textbf{Output}: answer $y$, trace $\tau$, updated library $\mathcal{R}_L'$
\begin{algorithmic}[1]
\STATE $\mathcal{C}\gets \textsc{Architect}(x)$; $\mathcal{C}\gets \textsc{FilterValid}(\mathcal{C})$
\STATE $\Delta\mathcal{R}\gets \textsc{SelectNovel}(\mathcal{C},\mathcal{R}_L)$; $\mathcal{V}\gets \mathcal{R}_L\cup\Delta\mathcal{R}$
\STATE $\mathcal{V}_K\gets \textsc{EpsGreedySelect}(\mathcal{V};\mathbf{w}_{\text{role}},K,\epsilon)$
\STATE $G_{\text{init}}\gets G_{\text{skel}}\ \cup\ \textsc{AddEdges}(\mathcal{V}_K;\mathbf{w}_{\text{edge}},\delta)$
\STATE $G_{\text{init}}\gets \textsc{EnforceDAG}(G_{\text{init}})$

\FOR{$t=1$ \textbf{to} $T_{\max}$}
    \STATE $(\tau,y)\gets \textsc{Execute}(G_{\text{init}},x)$
    \STATE $\mathcal{F}\gets \textsc{Feedback}(\tau)$; $p\gets \textsc{Pass}(\mathcal{F})$
    \IF{$p=1$} \STATE \textbf{break} \ENDIF
    \STATE $\mathcal{V}\gets \textsc{PromptRewrite}(\mathcal{V},\mathcal{F})$
    \STATE $G_{\text{init}}\gets \textsc{PriorFilteredExplore}(G_{\text{init}},\mathcal{F};\mathbf{w})$
\ENDFOR

\STATE $R\gets p-\lambda_{\text{cost}}\cdot \textsc{TokenCost}(\tau)$
\STATE $\mathbf{w}\gets \textsc{UpdatePriors}(\mathbf{w},R,\tau)$
\STATE $\mathcal{R}_L'\gets \mathcal{R}_L$
\IF{$p=1$}
    \STATE $\mathcal{R}_L'\gets \textsc{SolidifyTopK}(\mathcal{R}_L,\tau)$
\ENDIF
\STATE \textbf{return} $y,\tau,\mathcal{R}_L'$
\end{algorithmic}
\end{algorithm}

\paragraph{Verified Role Solidification and Reuse.}
In addition to updating priors, MetaGen maintains a growing pool of reusable roles.
During intra-task evolution, the Architect may synthesize new roles or substantially rewrite prompts to better match the instance.
To retain effective transient roles, we solidify roles only when the final execution passes task-specific checks.
Concretely, we extract a small Top-$K$ set of effective non-builtin roles from the executed graph (after de-duplication and basic validity checks), serialize their specifications, and store them in a persistent Role Cache.
In subsequent instances, the cache is loaded and merged into the role library as a high-priority candidate pool, enabling reuse of verified role templates rather than regenerating them from scratch.
Over time, this reward-conditioned retention expands the role library with task-relevant specialists and improves cold-start behavior under recurring patterns, without any backbone fine-tuning.

\begin{table}[t]
\centering
\small
\setlength{\tabcolsep}{6pt}

% single-column table (i.e., half of a two-column page width)
\resizebox{\columnwidth}{!}{%
\begin{tabular}{lccc}
\toprule
\textbf{Method} & \textbf{\#Training Token} & \textbf{\#Inference Token} & \textbf{\#Overall Token} \\
\midrule
Complete   & --                 & $9.8\times 10^{6}$ & $9.8\times 10^{6}$ \\
DyLAN      & $9.6\times 10^{6}$ & $1.3\times 10^{7}$ & $2.2\times 10^{7}$ \\
GPTSwarm   & $5.5\times 10^{6}$ & $8.4\times 10^{6}$ & $1.4\times 10^{7}$ \\
G-Designer & $2.7\times 10^{5}$  & $8.2\times 10^{6}$ & $8.5\times 10^{6}$ \\
\midrule
MetaGen    & -- & \textbf{$1.2\times 10^{6}$} & \textbf{$1.2\times 10^{6}$} \\
\bottomrule
\end{tabular}%
}

\caption{Token cost comparison measured with GPT-4.}
\label{tab:token_cost_gpt4}
\end{table}

\begin{figure}[t]
    \centering
    \begin{minipage}[t]{0.485\columnwidth}
        \centering
        \includegraphics[width=\linewidth]{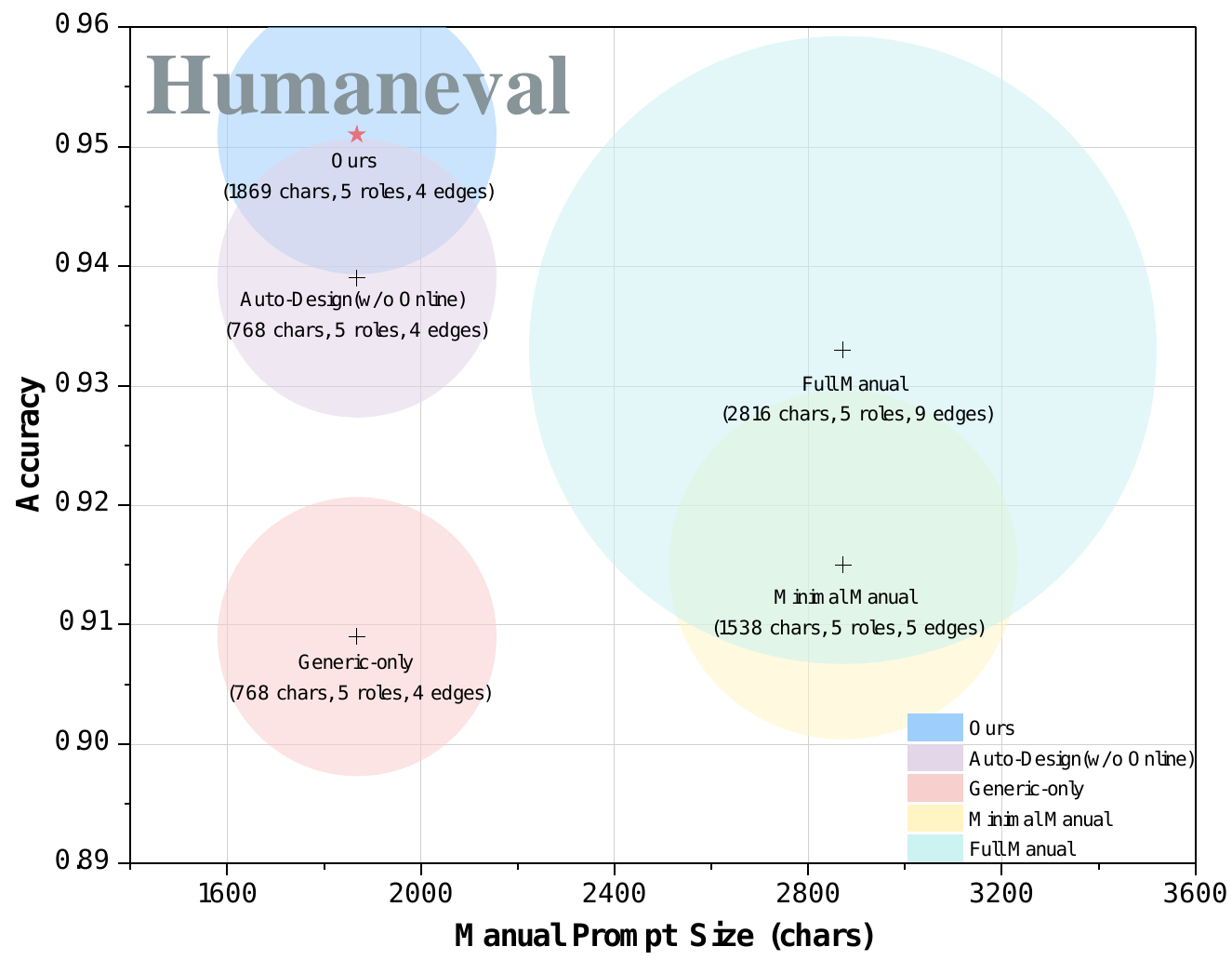}
    \end{minipage}\hfill
    \begin{minipage}[t]{0.485\columnwidth}
        \centering
        \includegraphics[width=\linewidth]{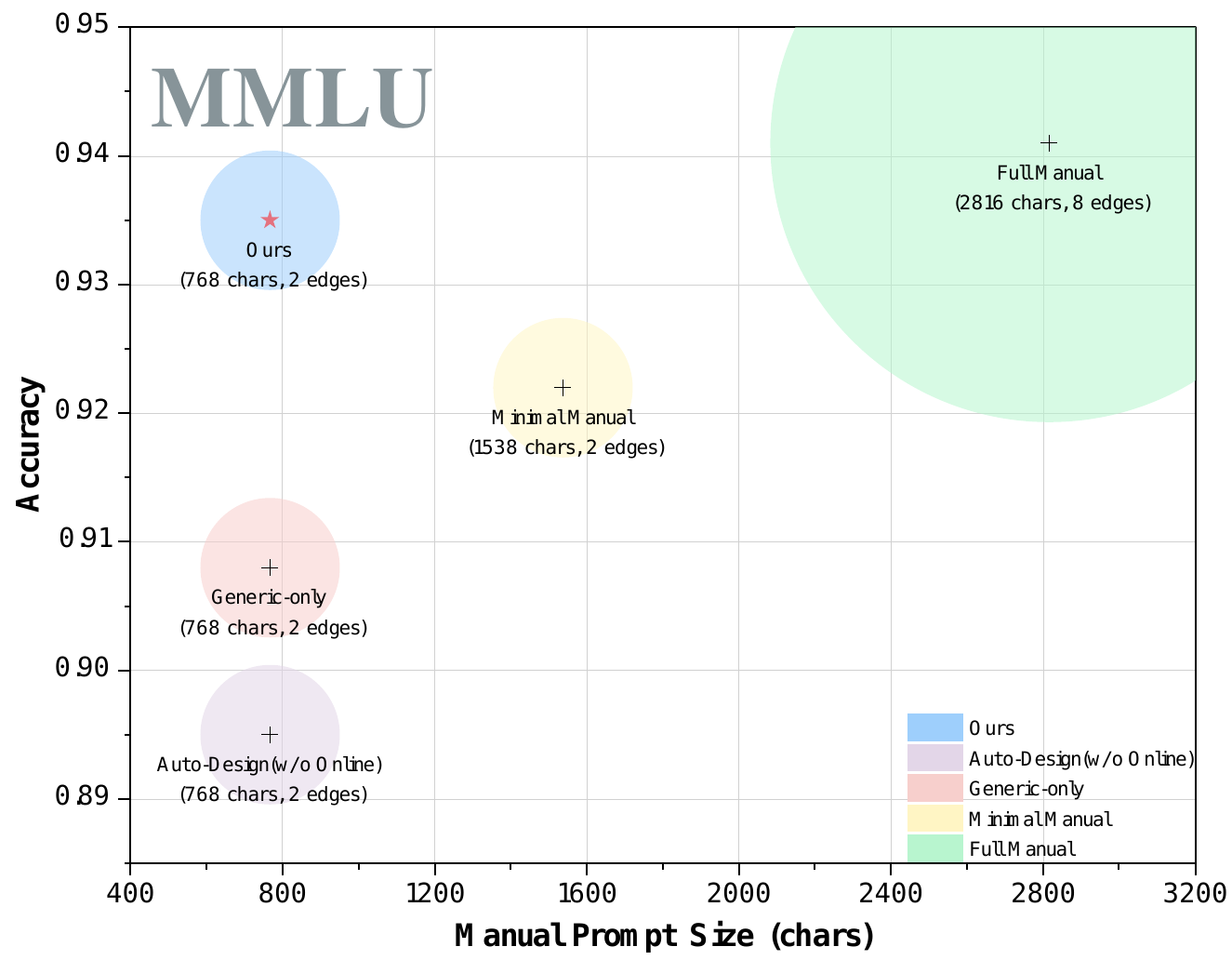}
    \end{minipage}
    \caption{Accuracy versus manual prompt size on HumanEval (left) and MMLU (right).
    Each point corresponds to a different design budget variant, illustrating the trade-off between engineering effort and performance.}
    \label{fig:manual_tradeoff}
\end{figure}

\section{Experiments and Analyses}

\subsection{Experimental Setup}
\label{sec:exp_setup}
\paragraph{Datasets and Metrics.}
We evaluate MetaGen on five widely used benchmarks that cover multi-step mathematical reasoning (GSM8K~\cite{gsm8k}), code generation (HumanEval~\cite{humaneval}), broad knowledge and reasoning (MMLU~\cite{mmlu}), algebraic word problems (AQuA~\cite{aqua}), and natural language inference (MNLI~\cite{mnli}).
For each dataset, we follow the official evaluation split and report the standard accuracy-based metric: exact-match accuracy for GSM8K and AQuA, classification accuracy for MMLU and MNLI, and pass@1 for HumanEval under the provided unit tests.
Our main comparison is summarized in Table~\ref{main}, where we also report the average score across the five datasets.

\paragraph{Baselines.}
We compare against both single-agent prompting and multi-agent orchestration baselines.
For single-agent prompting, we include a vanilla prompt, zero-shot and few-shot Chain-of-Thought\cite{CoT} prompting, and Self-Consistency\cite{selfcons} with multiple sampled rationales.
For fixed-topology multi-agent baselines, we instantiate common communication patterns, including Chain, Star, Tree\cite{qian2024scaling}, Complete Graph, Random Graph, as well as an LLM debate-style protocol\cite{debate}.
We further compare with representative automated topology design and multi-agent frameworks, including GPTSwarm\cite{gptswarm}, AFlow\cite{zhang2024aflow}, G-Designer\cite{g-designer}, and ARG-Designer\cite{arg-designer}.
When a baseline does not support a dataset in its original setting or public implementation, we mark the corresponding entry as missing in Table~\ref{main}.
%细节得改wang2020minilmdeepselfattentiondistillation
\paragraph{Implementation Details.}
All methods use the same backbone model, DeepSeek-V3~\cite{liu2024deepseek}, to isolate the effect of role generation and topology control. For the semantic encoder used in role relevance scoring and diversity control, we use SentenceTransformer \texttt{all-MiniLM-L6-v2}\cite{wang2020minilm}.
Unless otherwise specified, the Architect generates three candidate roles per instance and the selector instantiates a top-$K$ committee with $K{=}2$.
For online decision making, we use an $\epsilon$-greedy exploration strategy with $\epsilon{=}0.15$ and update step size $\eta{=}0.15$.
The reward trades off task success and cost as $R = \mathbb{I}(\text{pass}) - \lambda_{\text{cost}} \cdot \mathcal{C}_{\text{token}}$ with $\lambda_{\text{cost}}{=}0.001$.

\begin{table*}[t]
\centering
% 适当增加行间距，避免拥挤
\renewcommand{\arraystretch}{1.1} 

\resizebox{\textwidth}{!}{%
% 1. 去掉竖线，改为 lcccccccc
\begin{tabular}{lcccccccc}
\toprule
\multirow{2}{*}{\textbf{Method}} &
\multicolumn{2}{c}{\textbf{Overall 1--150}} &
\multicolumn{2}{c}{\textbf{Segment 1 MMLU 1--50}} &
\multicolumn{2}{c}{\textbf{Segment 2 MNLI 51--100}} &
\multicolumn{2}{c}{\textbf{Segment 3 HumanEval 101--150}} \\

% 2. 关键修改：添加 cmidrule 来区分不同的 Segment 组
% (lr) 参数让线条左右收缩，产生视觉间隔，非常美观
\cmidrule(lr){2-3} \cmidrule(lr){4-5} \cmidrule(lr){6-7} \cmidrule(lr){8-9}

& \textbf{Acc} & \textbf{AvgTok} &
  \textbf{Acc} & \textbf{AvgTok} &
  \textbf{Acc} & \textbf{AvgTok} &
  \textbf{Acc} & \textbf{AvgTok} \\
\midrule
Frozen & 90.0\% & 2673 & 92.0\% & \textbf{3030} & 80.0\% & 2782 & 94.0\% & 2208 \\
Random & 90.7\% & 2787 & 90.0\% & 3110 & 84.0\% & 3043 & 96.0\% & 2207 \\
\textbf{MetaGen} & \textbf{92.7\%} & \textbf{2483} & \textbf{94.0\%} & 3062 & \textbf{88.0\%} & \textbf{2190} & \textbf{100.0\%} & \textbf{2196} \\
\bottomrule
\end{tabular}%
}

\caption{Non-stationary stream evaluation on a 150-instance sequence.
The stream proceeds from Segment 1 MMLU, to Segment 2 MNLI, and then Segment 3 HumanEval.
We report accuracy and average total tokens per question for the full stream and each segment.}
\label{tab:nonstationary_overall_segments}
\end{table*}

\subsection{Main Results}
\label{sec:main_results}

Table~\ref{main} shows that MetaGen delivers the best overall performance, with a 1.8\% $\uparrow$ average accuracy improvement over the strongest baseline G-Designer.
On AQuA, MetaGen exceeds ARG-Designer by 5.1\% $\uparrow$, demonstrating clear benefits from role adaptation and inference-time topology evolution on multi-step reasoning.
On MNLI, MetaGen improves over the best reported baseline CoT few-shot by 9.4\% $\uparrow$, indicating substantially stronger task-conditional collaboration on NLI.
On HumanEval, MetaGen surpasses G-Designer by 0.9\% $\uparrow$, and on GSM8K it provides a further 0.1\% $\uparrow$ gain, while matching the best MMLU result.

\begin{figure}[t]
    \centering
    \includegraphics[width=\columnwidth]{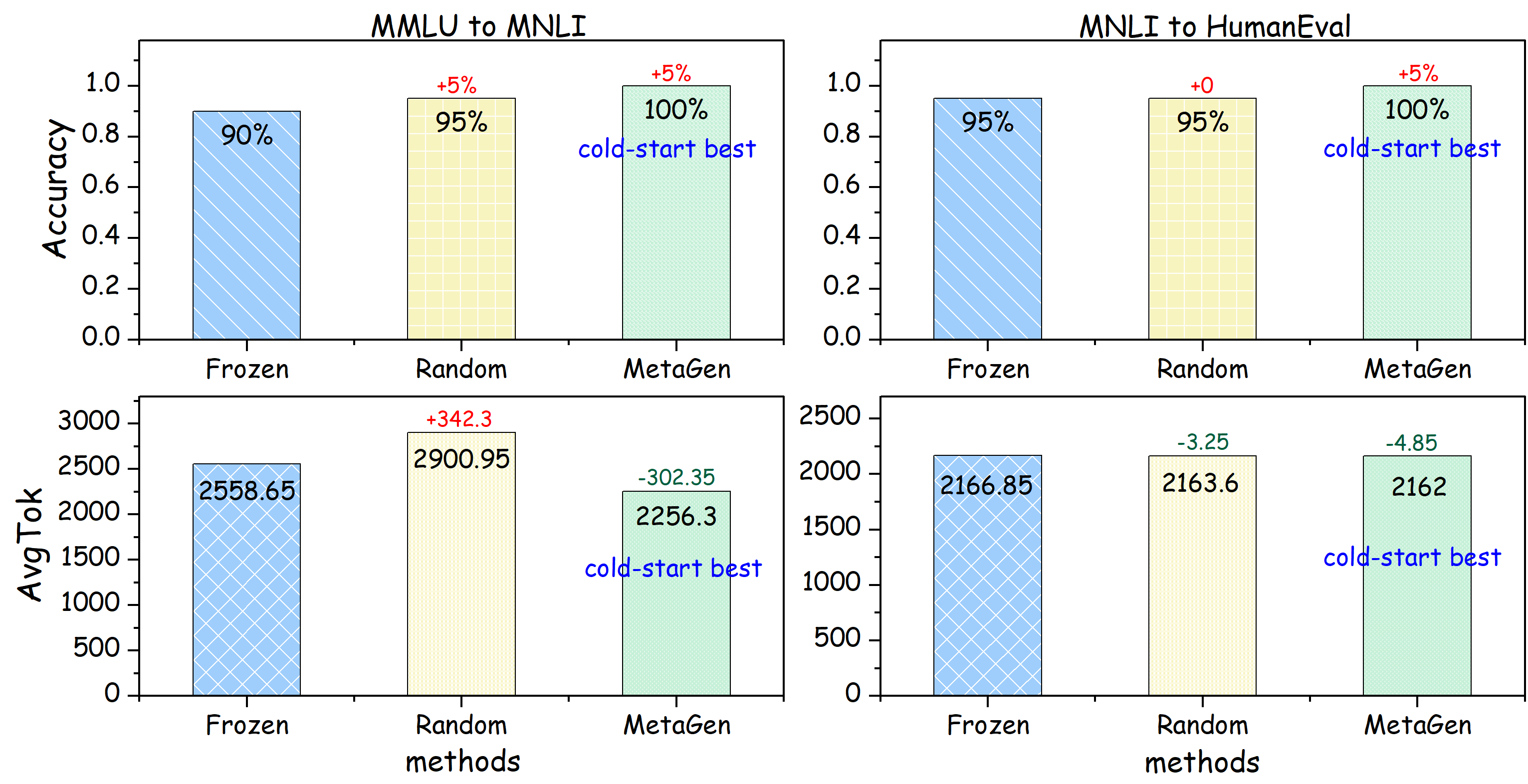}
    \caption{Cold-start recovery after distribution shifts.
    Accuracy (top) and average tokens (bottom) on the first 20 examples immediately after each shift, comparing Frozen, Random, and MetaGen.
    MetaGen achieves the strongest cold-start accuracy with lower token cost.}
    \label{fig:nonstationary_shift_post20}
\end{figure}

\subsection{Cost Efficiency}
\label{sec:cost_efficiency}

We evaluate cost from two complementary perspectives, runtime token cost and human authoring cost.

\paragraph{Runtime token cost.}
Table~\ref{tab:token_cost_gpt4} shows that MetaGen uses $1.2\times 10^{6}$ inference tokens.
This yields an 85.4\% reduction relative to G-Designer, an 87.8\% reduction relative to Complete, an 85.7\% reduction relative to GPTSwarm, and a 90.8\% reduction relative to DyLAN.
Overall, MetaGen achieves 7.1$\times$ to 18.3$\times$ fewer end-to-end tokens than prior multi-agent systems, while requiring no training tokens for role or topology design.

\paragraph{Human authoring cost.}
Figure~\ref{fig:manual_tradeoff} shows that MetaGen consistently improves the accuracy versus manual prompt size frontier on both HumanEval and MMLU.
At the same manual budget, MetaGen attains higher accuracy, and it reaches near-saturated performance with substantially less hand-written specification.
The advantage is most pronounced in the low-budget regime, indicating that online selection and evolution are the key drivers that recover accuracy when manual engineering is limited.

\begin{figure}[t]
    \centering
    \begin{minipage}[t]{0.485\columnwidth}
        \centering
        \includegraphics[width=\linewidth]{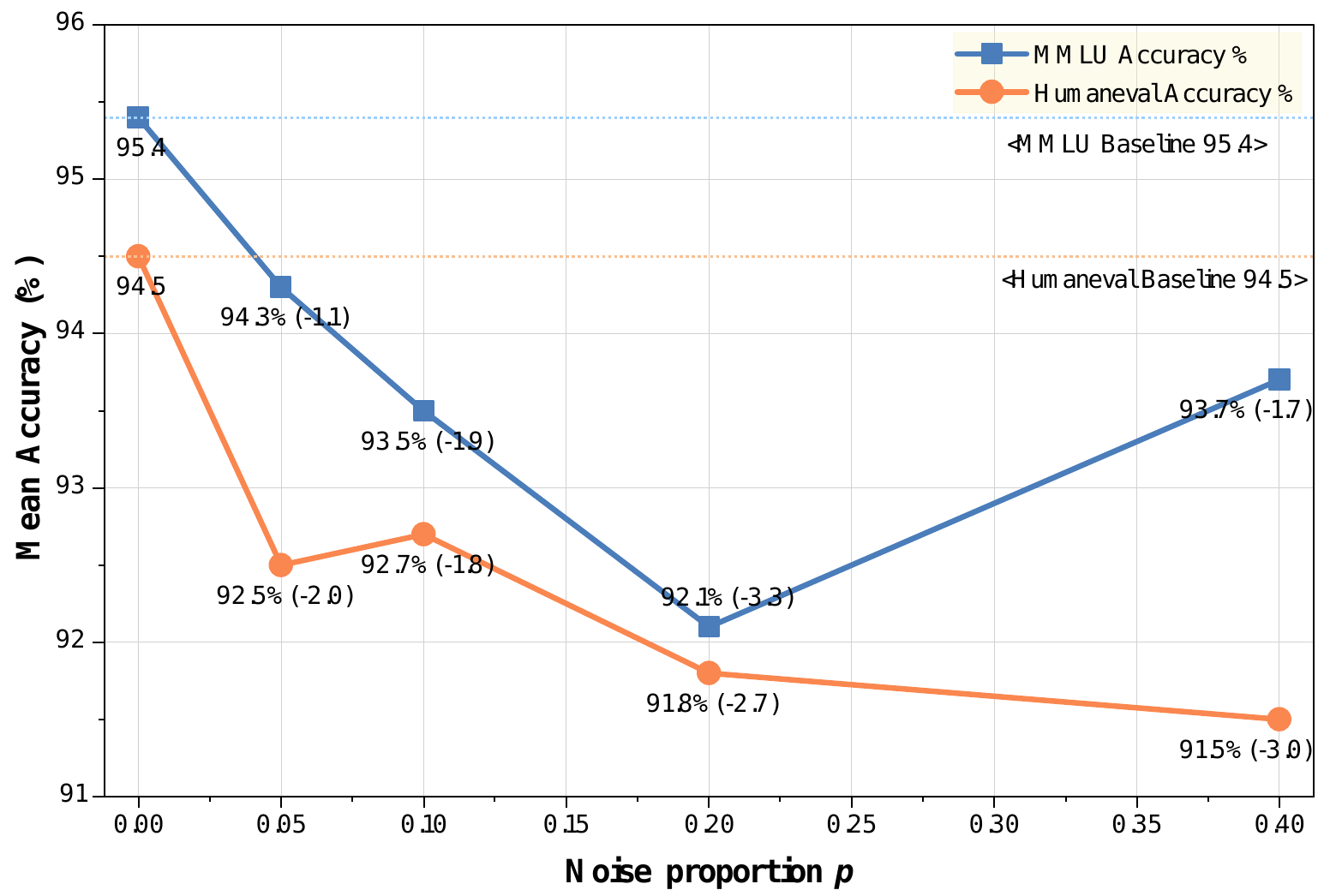}
    \end{minipage}\hfill
    \begin{minipage}[t]{0.485\columnwidth}
        \centering
        \includegraphics[width=\linewidth]{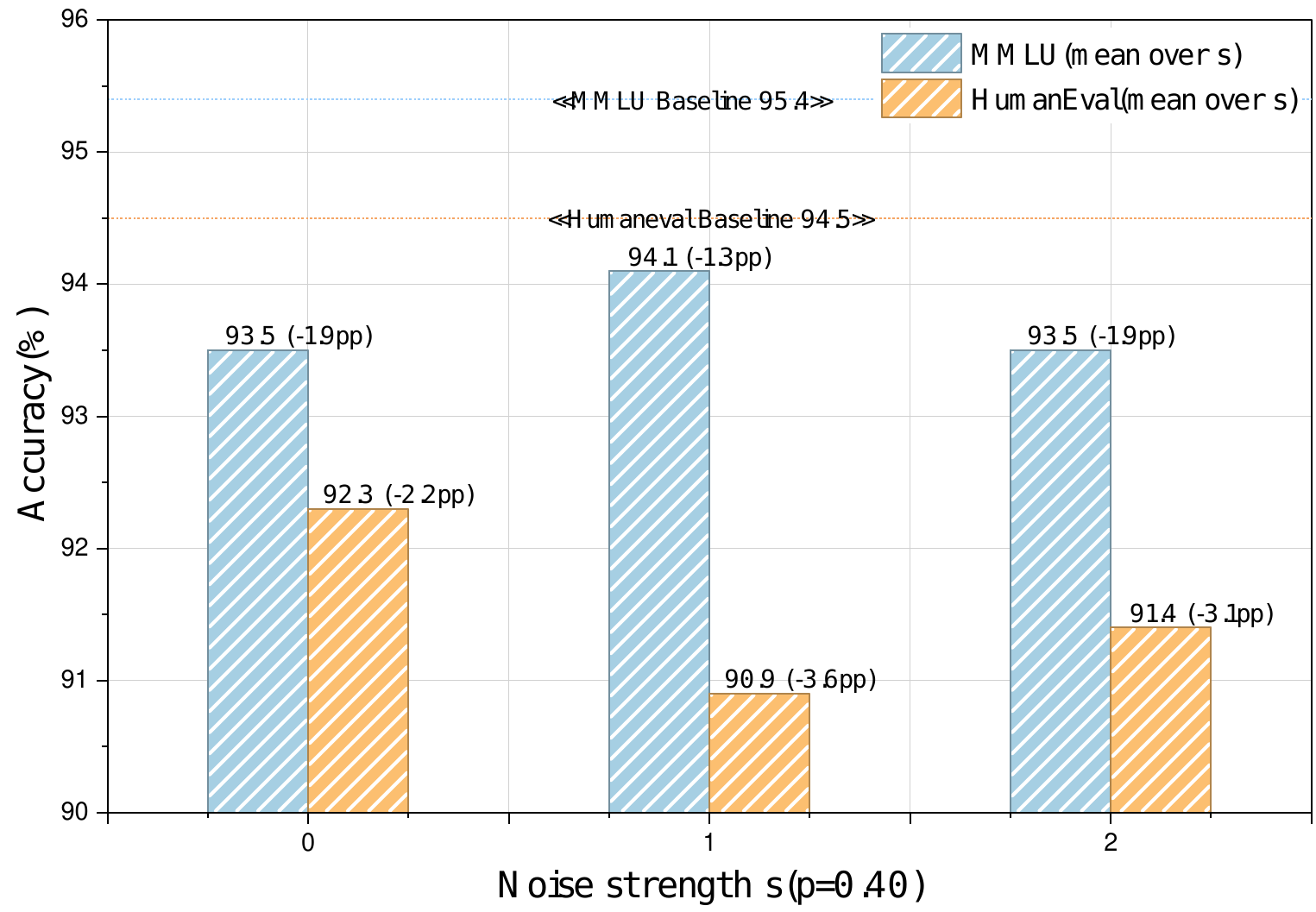}
    \end{minipage}
    \caption{Robustness to noisy nodes and edges.
    Left: varying the noise proportion $p$ (fraction of corrupted nodes and optional edges).
    Right: varying the noise strength level $s$ with fixed $p{=}0.4$.}
    \label{fig:noise_injection}
\end{figure}

\subsection{Adaptation and Robustness}
\paragraph{Non-stationary Stream Adaptation.}
We evaluate non-stationary adaptation on a 150-instance stream with three consecutive segments, consisting of 50 examples from MMLU, then 50 from MNLI, and finally 50 from HumanEval.
We compare MetaGen with Frozen, which keeps roles and topology fixed across the stream, and Random, which perturbs topology without learning.
Table~\ref{tab:nonstationary_overall_segments} shows that MetaGen improves overall accuracy by 3 points over Frozen and by 2 points over Random, while reducing average tokens by 7.1\% and 10.9\%.
The advantage is most pronounced on the shifted MNLI segment, where MetaGen improves accuracy by 8 points over Frozen and by 4 points over Random, with 21.3\% to 28.0\% fewer tokens.
On the final HumanEval segment, MetaGen reaches perfect accuracy, improving by 6 points over Frozen and by 4 points over Random without increasing tokens.

\paragraph{Cold-start Recovery After Distribution Shifts.}
Figure~\ref{fig:nonstationary_shift_post20} further isolates the first 20 examples immediately after each distribution shift.
After the first shift, MetaGen improves cold-start accuracy by 10 points over Frozen and by 5 points over Random, while reducing tokens by 11.8\% and 22.2\%.
After the second shift, MetaGen maintains perfect cold-start accuracy with no extra token overhead.
These results indicate that MetaGen not only adapts over the stream but also exhibits strong cold-start capability right after shifts.

\paragraph{Noise robustness.}
\label{sec:robustness_noise}
We test robustness by injecting corruption into both agent nodes and optional communication edges, controlled by a corruption ratio $p$ and a corruption strength $s$.
Figure~\ref{fig:noise_injection} shows that MetaGen remains stable under both widespread and stronger perturbations.
As $p$ increases, accuracy decreases gradually rather than collapsing, indicating that MetaGen does not hinge on any single critical node or edge and can preserve performance through redundant reasoning routes.
Notably, the degradation remains limited even at high corruption levels, suggesting that the evolving collaboration structure can compensate for partial failures by re-weighting or bypassing unreliable components.
When fixing $p{=}0.4$ and increasing $s$, performance exhibits only mild additional drops, implying that the system is resilient not only to the amount of noise but also to its severity.

\subsection{Ablation Study}
\label{sec:ablation}
We evaluate four MetaGen variants by disabling one mechanism at a time. \textbf{(1) w/o Role Generation} uses a fixed role set without query-conditioned role synthesis. \textbf{(2) w/o Learned Policy} replaces learning-based selection and wiring with random or relevance-only heuristics, and does not use persistent statistics or policy states for decision making. \textbf{(3) w/o Intra-task Evolution} disables within-instance updates so the system executes $G_{\text{init}}$ without prompt rewriting or topology adjustment. \textbf{(4) w/o Cross-instance Memory} disables persistence across instances by stopping verified role write-back and resetting selection and wiring states so each instance cold-starts. Table~\ref{tab:ablation} shows that each component matters and the full MetaGen performs best. Removing role generation yields the most pronounced degradation, for example dropping HumanEval from 95.1 to 92.1, which highlights the importance of query-conditioned role instantiation. Replacing the learned policy also reduces accuracy, such as to 92.8 on MMLU, indicating that learned decision rules for selecting participants and optional connections are beneficial beyond having a larger candidate set. Disabling intra-task evolution lowers performance to 91.7 on MMLU, suggesting that refining prompts and structure within an instance materially improves solution quality. Finally, removing cross-instance memory degrades results to 92.7 on HumanEval, showing that persistent accumulation of verified roles and selection statistics improves robustness across instances.

\begin{table}[t]
    \centering
    \scriptsize
    \setlength{\tabcolsep}{4.8pt}
    \renewcommand{\arraystretch}{1.08}
    \resizebox{\columnwidth}{!}{%
    \begin{tabular}{lcc}
        \toprule
        \textbf{Variant} & \textbf{HumanEval} & \textbf{MMLU} \\
        \midrule
        vanilla MetaGen  & 95.1 & 93.5 \\
        \midrule
        w/o Role Generation & 92.1\dec{3.0} & 91.1\dec{2.4} \\
        w/o Learned Policy   & 93.9\dec{1.2} & 92.8\dec{0.7} \\
        w/o Intra-task Evolution & 92.7\dec{2.4} & 91.7\dec{1.8} \\
        w/o Cross-instance Memory & 92.7\dec{2.4} & 92.6\dec{0.9} \\
        \bottomrule
    \end{tabular}%
    }
    \caption{Ablation study. Each variant removes one component from MetaGen.}
    \label{tab:ablation}
\end{table}

\section{Conclusion}
We propose \textbf{MetaGen}, a training-free multi-agent framework that improves accuracy while reducing both inference-token cost and manual prompt engineering by generating and refining roles and collaboration structure at inference time.
With a DeepSeek-V3 backbone, MetaGen achieves the strongest overall performance across five benchmarks against single-agent prompting, fixed-topology orchestration, and topology-design baselines.
Further analyses show that MetaGen degrades gracefully under noisy agents and perturbed edges, benefits from each core component, and adapts to non-stationary task streams with strong cold-start recovery after distribution shifts.
Overall, these results highlight inference-time optimization of text-level roles and discrete collaboration structure as a practical path toward scalable and adaptive MAS without modifying backbone weights.

\bibliographystyle{named}
\bibliography{ijcai26}

\end{document}